\documentclass[letterpaper, 10 pt, conference]{ieeeconf}  

\IEEEoverridecommandlockouts                              

\usepackage[caption=false]{subfig}
\usepackage{graphics} 
\usepackage{graphicx}

\usepackage{epsfig} 
\usepackage{amsmath} 
\usepackage{amssymb}  
 \usepackage{algorithm}
\usepackage[compatible]{algpseudocode}
\algnewcommand\AAND{\textbf{ and }}
\algnewcommand\Or{\textbf{ or }}
\usepackage{color}
\usepackage{citesort}
\usepackage{flushend}
\usepackage{url}
\usepackage[table,xcdraw]{xcolor}
\usepackage{amsmath}

\usepackage[colorinlistoftodos]{todonotes}

\DeclareMathAlphabet{\pazocal}{OMS}{zplm}{m}{n}

\DeclareMathAlphabet{\mathpzc}{OT1}{pzc}{m}{it}

\usepackage{array}
\newcolumntype{C}[1]{>{\centering\arraybackslash}p{#1}}
\newcolumntype{M}[1]{>{\raggedright\arraybackslash}p{#1}}

\usepackage{array} 
\newcolumntype{L}[1]{>{\raggedright\let\newline\\\arraybackslash\hspace{0pt}}m{#1}}	
\newcolumntype{S}[1]{>{\centering\let\newline\\\arraybackslash\hspace{0pt}}m{#1}}
\newcolumntype{R}[1]{>{\raggedleft\let\newline\\\arraybackslash\hspace{0pt}}m{#1}}

\usepackage[nolist,nohyperlinks]{acronym}
\acrodef{mlt}[MLT]{Martian Lava Tube}
\acrodef{drl}[DRL]{Deep Reinforcement Learning}
\acrodef{ppo}[PPO]{Proximal Policy Optimization}

\makeatletter
\renewcommand*{\@opargbegintheorem}[3]{\trivlist
  \item[\hskip \labelsep{\itshape #1\ #2}] \textit{(#3)}\ }
\makeatother

\title{\LARGE \bf
 Olympus: A Jumping Quadruped for Planetary Exploration Utilizing Reinforcement Learning for In-flight Attitude Control
}

\author{J{\o}rgen Anker Olsen, Grzegorz Malczyk, and  Kostas Alexis
\thanks{
}%
\thanks{This material was supported by the Research Council of Norway under Award NO-338694. The authors are with the Autonomous Robots Lab, NTNU, O.S. Bragstads Plass 2D, 7034, Trondheim, Norway. {\tt\small jorgen.a.olsen@ntnu.no} \newline
}
}

\begin{document}

\maketitle
\thispagestyle{empty}
\pagestyle{empty}

\begin{abstract}
Exploring planetary bodies with lower gravity, such as the moon and Mars, allows legged robots to utilize jumping as an efficient form of locomotion thus giving them a valuable advantage over traditional rovers for exploration. Motivated by this fact, this paper presents the design, simulation, and learning-based ``in-flight'' attitude control of Olympus, a jumping legged robot tailored to the gravity of Mars. First, the design requirements are outlined followed by detailing how simulation enabled optimizing the robot's design - from its legs to the overall configuration - towards high vertical jumping, forward jumping distance, and in-flight attitude reorientation. Subsequently, the reinforcement learning policy used to track desired in-flight attitude maneuvers is presented. Successfully crossing the sim2real gap, extensive experimental studies of attitude reorientation tests are demonstrated.

\end{abstract}

\section{INTRODUCTION}\label{sec:intro}

Extraterrestrial robotic missions have a long history and have been the prime strategy for space exploration during the last decades. Predominantly, this involves large, highly competent, multi-purpose, high-cost rovers deployed to explore the surface of Earth's planetary neighbors. More recently, we have seen the utilization of flying systems, while currently multiple alternative concepts are being considered to drive the future of space exploration. Motivated by their locomotion dexterity and successful deployments in challenging terrestrial environments, including underground caves and mines~\cite{tranzatto2022cerberus,agha2021nebula}, legged robots have recently attracted attention as a promising approach to planetary exploration missions. This is especially the case within the context of deployments that will have to deal with perilous terrain such as within the complex underground networks of \acp{mlt}~\cite{crown2022distribution}. \acp{mlt} have attracted the attention of the scientific community owing to their potential to offer access to valuable resources, a window to the possible biological history of the planet, as well as an environment for possible human settlements protected from radiation and other challenges present on the surface of Mars~\cite{leveille2010lava,sauro2020lava,perkins2020lava,mari2021potential}.

\begin{figure}
    \centering
    \includegraphics[width=0.46\textwidth]{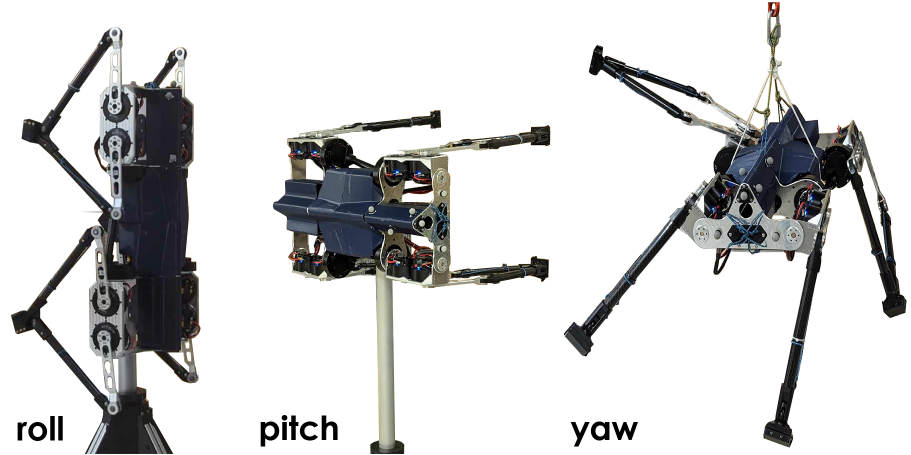}    \vspace{-2ex}
    \caption{The Olympus quadruped mounted for experimental setups for reorientation tests - roll evaluation on rotating rod (left), pitch assessment on rotating rod (center), and yaw test using fixed rope suspension (right). }
    \label{fig:front_figure}
    \vspace{-4ex}
\end{figure}

Nevertheless, the utilization of legged systems within the framework of such missions is anything but straightforward. Access to \acp{mlt} is available mostly through skylights and other narrow passages which implies that the employed systems have to remain small in size. In addition, their complex multi-branching geometry may imply that a single larger system could perform worse than a team of smaller systems that otherwise correspond to the same overall weight and volume (two commanding parameters defining costs and feasibility of a mission to Mars). Considering the small size with the likelihood of highly complex and anomalous terrain inside \acp{mlt}, locomotion flexibility can be achieved by systems that exploit the reduced gravity of Mars (approximately a third of that on Earth) to perform long jumps, allowing them to overcome large obstacles and ditches.

Motivated by the above, in this work we present the design of a novel quadruped tailored to low-gravity jumping, alongside a \ac{drl}-based control policy for the in-flight attitude stabilization of the system. The robot, called ``Olympus'', employs a 5-bar mechanism for the design of each of its legs with the link ratios and the internally integrated spring optimized for high jumping on Mars~\cite{olsen2023design}, and the overall configuration tailored to agile reorientation maneuvering. Exploiting fast and precise motor control, the large workspace of each of the legs' kinematics, and the relative placement of front and back legs reduces the likelihood of self-collisions. The \ac{drl} policy is trained to stabilize the robot's attitude and track desired maneuvers based on \ac{ppo}, successfully crossing the sim2real gap. The approach is not only evaluated in simulation, but also in detailed experimental studies that involve Olympus performing a family of attitude maneuvers.

The remainder of this paper is organized as follows: Section~\ref{sec:related} presents related work, followed by the system description along with simulation-based design optimization detailed in Section~\ref{sec:sys_des}. The RL problem and training pipeline are presented in Section~~\ref{sec:RL_attitude}, while simulation and experimental results are detailed in Section~\ref{sec:_Eval_Intro}. Finally, conclusions are drawn in Section~\ref{sec:conclusions}.

\section{RELATED WORK}\label{sec:related}
This work relates to three bodies of literature, namely that of a) novel robot concepts considered for traversing the challenging terrain of martian lava tubes, b) jumping legged robots, and c) control strategies to achieve the in-flight attitude stabilization of such systems. 

Space exploration missions have historically leveraged the power and comparative benefits (e.g., in human safety) of robot deployments. Concepts regarding future space missions consider a variety of designs to undertake strenuous challenges such as that of navigating the perilous terrain of Martian lava tubes. Examples include legged robots~\cite{morrell2024robotic,arm2023scientific,olsen2023martian}, multirotor flying systems~\cite{patel2023towards}, hopping-and-flying vehicles~\cite{thangavelautham2017flying}, inflatable platforms~\cite{dinkel2024martian}, or unconventional concepts such as the ReachBot design~\cite{newdick2023designing}. Within the multitude of approaches, quadrupeds have recently attracted significant attention primarily driven by the overall progress of the field in dexterous locomotion~\cite{rudin2022learning,chai2022survey}, and the accelerating effects induced by the DARPA Subterranean Challenge~\cite{tranzatto2022cerberus}. 

Recently, a niche set of works has focused on high-jumping quadrupeds in low-gravity environments. Equipped with such a capacity, future quadrupeds may be able to exploit jumping to seamlessly traverse the complex terrain of \acp{mlt} especially when it comes to negotiating obstacles and ditches much larger than their own robot size. To that end, a set of works has looked into specialized system design and control policy derivation. The works in~\cite{Spacebok1,Spacebok2} introduced Spacebok, a system with legs relying on a parallel mechanism and capable of both jumping and regulating its attitude mid-flight. \ac{drl} was utilized in~\cite{Spacebok2} to achieve jumping and safe landing, albeit in simulation. The SpaceHopper system was introduced in~\cite{spiridonov2024spacehopper} and represents a three-legged system tailored for exploration of asteroids and moons having very low gravity conditions. The authors' previous work in~\cite{olsen2023martian,olsen2023design} outlined a concept design for a jumping quadruped tailored to \ac{mlt} navigation and described its leg design optimized for jumping capacity in Mars' gravity. More broadly, works on such low-gravity jumping explorers relates to the overall body of work on jumping and in-flight attitude stabilizing such as the works in~\cite{Spacebok1,wang2023research,bellegarda2024quadruped,kurtz2022mini}. Focusing on landing on asteroids, the authors in~\cite{qi2023integrated} present a learning-based reorientation strategy with results only in simulation. Compared to existing literature, this work builds upon the core ideas in~\cite{Spacebok1,Spacebok2,olsen2023martian} and presents ``Olympus'', a quadruped robot the design of which is optimized for high-power jumping on Mars, alongside its control policy for in-flight attitude maneuvering trained through \ac{drl}.

\section{SYSTEM DESIGN}\label{sec:sys_des}
This section outlines the robot design, body, and leg optimization. 

\subsection{Design goals, constraints, and operational assumptions}
The proposed robot, Olympus, is designed to meet the unique challenges of exploring Martian lava tubes. Its development is guided by a set of carefully considered goals, constraints, and requirements that shall enable the robot to explore both on the surface and subterranean of lower-gravity planetary bodies and especially Mars. Key design goals for the robot to achieve this are: a) Capability to make large vertical and forward jumps to overcome obstacles or ditches larger than itself,  b) a significant workspace that allows for attitude corrections and reorientation post jump, in the flight phase, to maintain an orientation that results in a safe landing, c) the ability for robust walking locomotion and smaller jumps for locomotion.
The primary constraints for Olympus focus on minimizing mass and maximizing jump height in the Martian gravity environment ($\approx$0.38~\textrm{g}). To achieve these design targets, a quadruped configuration was selected, driven by the dexterous locomotion capabilities of such systems. The actuation, leg design, compliant elements, and overall kinematic and mechatronic configuration of the system represent the design space that needs to be optimized. 

\subsection{Leg Design}
Key factors to maximize the jump height are a) the leg kinematic design, b) the use of fast and high-torque actuators, in combination with c) springs and elastic/compressible components. Modern legged systems utilize legs with various designs with the most common being the open kinematic chain found in ANYbotics ANYmal \cite{hutter_anymal_2016}, Unitree A1 \cite{unitree_a1_website}, and Boston Dynamics Spot \cite{spot_website}. Nevertheless, some systems utilize less common designs that can enhance jumping capabilities, such as the closed kinematic designs of Minotaur~\cite{kenneally_design_2016} and SpaceBok~\cite{arm_spacebok_2019}.

For Olympus the 5-bar closed kinematic chain was selected for the leg design due to the advantages it provides in a) enabling two motors to contribute fully to applying the downward force needed for jumping, as well as b) allowing to integrate springs between two of the hip links to aid in jumping and energy recovery while landing \cite{kenneally_leg_2015}. This, combined with a motor for hip actuation, simultaneously provides a large workspace to move the legs, both for walking and in-flight reorientation \cite{olsen2023design}. The optimization of the 5-bar link lengths is detailed in Section~\ref{sec:designopt}. For this prototype, factors such as power consumption, communication limitations, dust mitigation, and radiation resistance are not considered more than in \cite{olsen2023martian}, allowing focus on the core jumping locomotion and reorientation capabilities.

\subsection{Electrical system and actuation control}\label{sec:electricalsys}
The Olympus prototype demands high-torque, rapid actuators for optimal jumping and reorientation. We selected two off-the-shelf CubeMars models: the AK70-10 and AK80-9, offering a balance of affordability, torque, and speed. The AK70-10 delivers a peak torque of $24.8$~\textrm{Nm} at a $23.2$~\textrm{A} peak current, a top speed of $310$~\textrm{RPM}, and weighs $610$~\textrm{g}, utilizing a $10$:$1$ planetary gear. In contrast, the AK80-9 provides $18$~\textrm{Nm} peak torque at a $22.3$~\textrm{A} peak current, $390$~\textrm{RPM}, and weighs $485$~\textrm{g}, with a $9$:$1$ planetary gear. The robot employs AK70-10 actuators for the 5-bar mechanism links, leveraging their higher torque for jumping capabilities. AK80-9 actuators are used for hip actuation to reduce weight. Both models communicate via CAN-Bus at speeds up to $1$~\textrm{MHz}. For high-rate data transmission and redundancy, four Innomaker USB2CAN modules are utilized, one per leg, connected to a USB hub. The onboard computer is an NVIDIA Jetson Orin NX $16$ GB system mounted on an A603 Carrier Board, which processes the CAN-Bus communication. A VectorNav VN-100 Inertial Measurement Unit (IMU) is further integrated.

Power to the robot is supplied by four Tattu R-Line V5.0 $6$S LiPo batteries ($1200$~\textrm{mAh}, $22.2$~\textrm{V}, $150$~\textrm{C}). Two batteries are installed in series to form a $12$S configuration ($44.4$~\textrm{V}, $150$~\textrm{C}) for each pair of legs. Two Blida XT30 Power Distribution Board (PDB) (XT90 input, $8 \times$ XT30 outputs) manages power delivery to the legs, while a DC-DC converter powers the onboard computer, which in turn powers the remaining electronics on the robot through USB. This high-current capacity setup accommodates a three-tier power consumption strategy: minimal during rest, low while walking, and peak during jumping, re-orientation, and landing.

The onboard computer executes a ROS-based motion control and the motor control system implemented in C++. The latter operates four independent ROS nodes, one per leg, communicating via CAN bus to each motor's driver board on that leg for redundancy. During reorientation tests, the motors' PID controller operates at a frequency of $200$~\textrm{Hz}.

\subsection{Robot Leg and Body Parameters Optimization}\label{sec:designopt}
Prior to robot construction, comprehensive simulation studies were conducted to optimize leg and body configurations. These simulations, performed through a parallel grid search of all parameters in MATLAB Simulink with Simscape multibody, aimed to achieve two primary objectives: maximizing jump height and enabling effective in-flight reorientation. The latter capability is crucial for correcting undesired attitude errors or angular rates imparted during jumps, ensuring safe landing orientations.

\begin{figure}
    \centering
    \includegraphics[width=0.8\columnwidth]{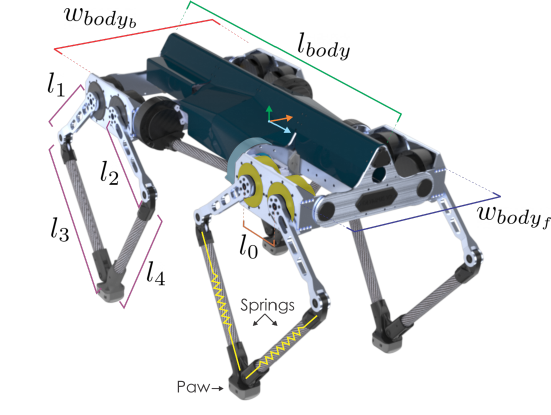}
    \vspace{-3ex}
    \caption{CAD render of Olympus, depicting a) the core component lengths of the robot, b) highlight of the $3$ motors of each leg for the front right motor (yellow used for the 5-bar motors and cyan for the hip motor), as well as c) where the springs are integrated within each leg.}
    \label{fig:explanation_figure}
    \vspace{-5ex}
\end{figure}

The robot body and leg links are illustrated in Figure~\ref{fig:explanation_figure}. The parameter space iterated for the robot leg included the link lengths $l_1$, $l_2$, $l_3$, and $l_4$ of the 5-bar mechanism, configured in a diamond design where $l_1 = l_2$ and $l_3 = l_4$. The spring stiffness $k_s$ was optimized and represents a spring connected via a cord in the joints between $l_1$-$l_3$ and $l_2$-$l_4$. A previous simulation study of the authors compared this diamond configuration with a parallelogram design ($l_1 = l_4$ and $l_2 = l_3$) and demonstrated superior jump height for the diamond design~\cite{olsen2023design}. For the robot body, parameters included body length $l_{body}$, front leg separation $w_{body_f}$, and back leg separation $w_{body_b}$. The range of values iterated through for the leg and body parameters are listed in Table \ref{tab:OLYMPUSsimulationparameters} and their appropriate placement on the robot in Figure \ref{fig:explanation_figure}. Note that for purposes of navigation symmetry, it was imposed that front and back legs are identical.

The simplified Simulink model has rectangular prisms representing the body (length $l_{body}$, front leg separation $w_{body_f}$, rear leg separation $w_{body_b}$) and the 5-bar motor mount. Links $l_1$ and $l_2$ were modeled as rectangular prisms, while $l_3$ and $l_4$ were represented by cylinders. A $0.05$~\textrm{m}-diameter sphere modeled the paw. The physical system modeling incorporated rotational joints at motor and axis locations. Simulations employed a variable-step solver with the \texttt{ode45} algorithm, optimizing computational efficiency and simulation duration. The natural length of the inter-knee joint spring was set to $0.18$~\textrm{m}, corresponding to the threshold at which the planned integrated springs would initiate actuation upon leg extension.

The simulations utilized the AK70-10 and AK80-9 motor parameters. Constant masses were maintained for peripheral components (mounting hardware, batteries, Orin processor, IMU, PDC, and USB2CAN devices). Link masses were derived from density-based estimates, correlating with material properties, geometry, and CAD-approximated dimensions. Body mass calculations initiated from a baseline configuration, incorporating i) motor masses at simulated revolute joints representing motor mounting sections for the legs, ii) a centralized block representing electronic components, and iii) estimated structural elements. Incremental increases in body size resulted in proportional mass augmentations, reflecting additional material requirements for larger systems. The constant length $l_0$ of the 5-bar leg  was set to $0.09$~\textrm{m}, corresponding to the minimum achievable distance between adjacent AK70-10 motor axes in the 5-bar configuration.

Simulations were conducted for vertical and horizontal jumps under Martian gravity ($3.721$~\textrm{m/s$^2$}), and roll, pitch, and yaw maneuvers in zero gravity. Vertical jump simulations assessed maximum height, while horizontal jump simulations evaluated distance, height, and orientation errors during flight caused by slight variations in individual leg take-off forces. Both jump types and attitude maneuvers were simulated for all parameter combinations for leg and body.

The grid search employed a proportional-integral-derivative (PID) controller commanding joint angles. The jump sequence was structured as follows: a) Initial position: All leg links vertical in zero position, paws grounded, b) Squat motion: $1$~\textrm{s} duration, motors 1 and 2 for each leg moving from $0$~\textrm{deg} to $120$~\textrm{deg} and $0$~\textrm{deg} to $-120$~\textrm{deg} respectively for $l_1$ and $l_2$, c) Hold: $0.5$~\textrm{s} and d) Jump initiation: Step input from $120$~\textrm{deg} to $10$~\textrm{deg}, utilizing maximum torque (saturated at $\tau_{max} = 24$~\textrm{Nm}). 
The hip motors maintained zero position throughout the simulated jumps. For horizontal jumps, an additional forward lean phase was introduced after the squat: a) Forward lean: $1$~\textrm{s} duration, jump motor setpoints shifted by $45$~\textrm{deg}, b) Hold leaned position: $0.5$~\textrm{s} and c) Jump initiation as in vertical jump.

Attitude reorientation tests evaluated the robot's attitude correction capabilities in roll, pitch, and yaw during a 5-second flight phase. Preprogrammed motion primitives governed the leg movements for each actuator. These primitives were designed to maximize leg workspace utilization across various simulated leg and body dimensions. The reorientation strategy employs single-axis leg actuations, alternating between maximum extension for high inertia and maximum retraction for low inertia, effecting attitude changes in zero-gravity conditions.

\begin{table}[t]
\begin{center}
\caption{Robot simulation parameters and their Search Space}
\label{tab:OLYMPUSsimulationparameters}
\begin{tabular}{l|c|c|c}
Parameter                    & Symbol        & Search Values                & Design\\ \hline
Mass leg                     & $m_l$         & $0.39-0.57$ {[}kg{]} & $0.63$ {[}kg{]}\\
Mass electronics             & $m_e$         & $2.1$ {[}kg{]}       & $2.0$ {[}kg{]}\\
Mass motors                  & $m_m$         & $6.82$ {[}kg{]}      & $6.82$ {[}kg{]}\\
Body length                  & $l_{body}$    & $0.4-1.0$ {[}m{]}    & $0.6$ {[}m{]}\\
Front leg separation         & $w_{body f}$  & $0.2-0.6$ {[}m{]}   & $0.21$ {[}m{]}\\
Back leg separation          & $w_{body b}$  & $0.2-0.6$ {[}m{]}    & $0.3 $ {[}m{]}\\
Link 0 length                & $l_0$         & $0.09$ {[}m{]}       & $0.09$ {[}m{]} \\
Link 1 length                & $l_1$         & $0.10-0.30$ {[}m{]}  & $0.175$ {[}m{]}\\
Link 2 length                & $l_2$         & $0.10-0.30$ {[}m{]}  & $0.175$ {[}m{]}\\
Link 3 length                & $l_3$         & $0.15-0.45$ {[}m{]}  & $0.3$ {[}m{]}\\
Link 4 length                & $l_4$         & $0.15-0.45$ {[}m{]}  & $0.3$ {[}m{]}\\
Spring stiffness             & $k$           & $600-1000$ {[}N/m{]} & $800$ {[}N/m{]}
\end{tabular}
\end{center}
\vspace{-4ex}
\end{table}

The grid search optimization employed a weighted scoring system to evaluate the multi-dimensional results. For each configuration, key performance metrics were recorded and normalized: a) Maximum vertical jump height $h_{max}$ adjusted for leg length, b) Maximum horizontal displacement $d_{max}$ and height at apogee $h_{y}$ in the x-y plane, c) Maximum body rotation rates $\omega_{roll}, \omega_{pitch}, \omega_{yaw}$ post-takeoff, and d) Maximum angles $\theta_{roll}, \theta_{pitch}, \theta_{yaw}$ achieved after $5$~\textrm{s} of applied motion primitives. The scoring weights were assigned as follows: vertical jump $h_{max}$: 4, horizontal jump $h_{y}$: 3 and $d_{max}$: 3, absolute pitch angle: -1, and reorientation angles $\theta_{roll}$: 2, $\theta_{pitch}$: 2), and $\theta_{yaw}$: 1. This approach facilitated effective comparison across diverse robot parameter combinations. 

\begin{figure}
    \centering
    \includegraphics[width=0.48\textwidth]{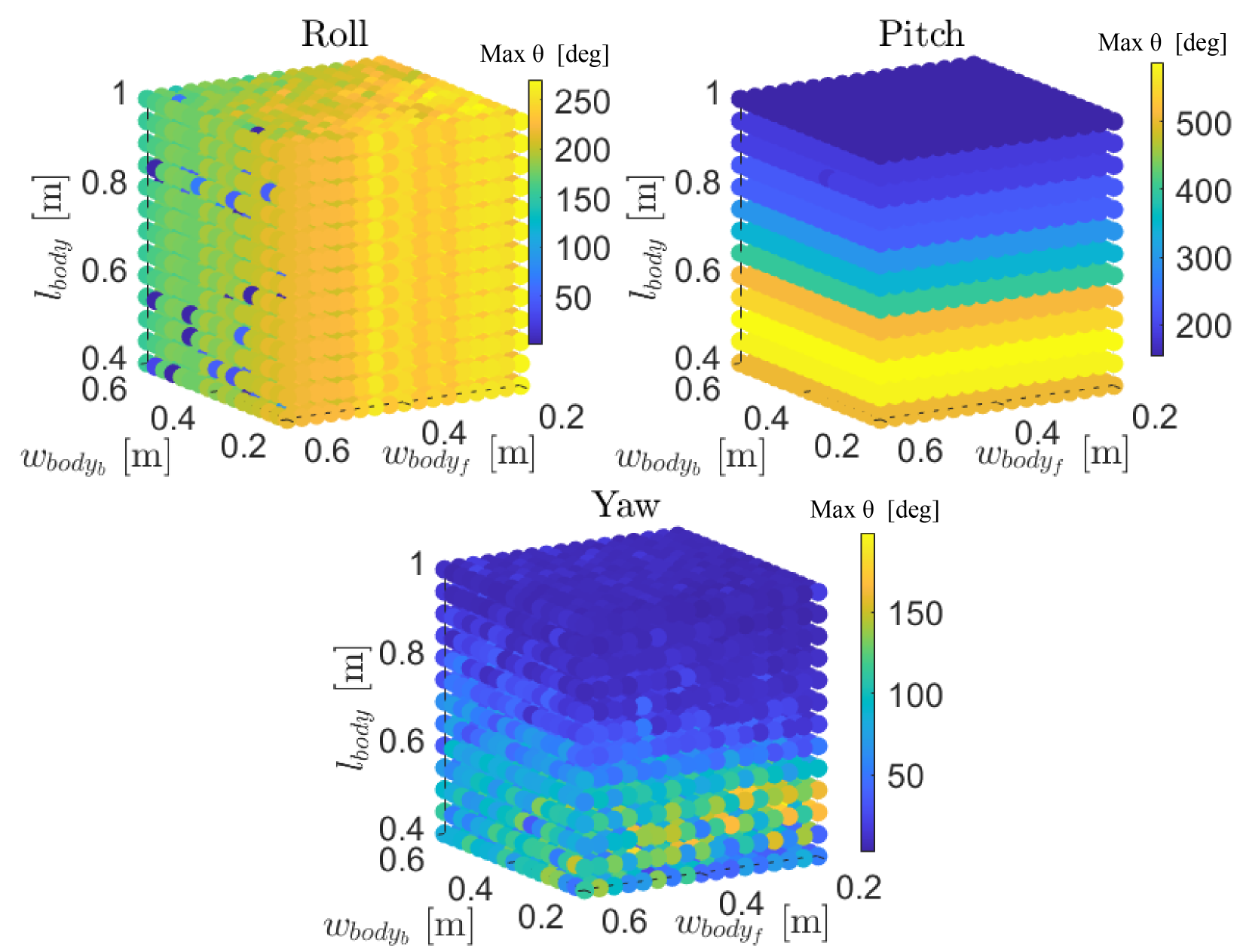}
    \caption{Parameter space exploration of body dimensions: body length $l_{body}$, front leg separation $w_{body f}$, and back leg separation $w_{body f}$. Heat map illustrates performance correlation with body size parameters for roll, pitch, and yaw reorientation in the grid search optimization, maintaining fixed link lengths.}
    \label{fig:grid_search_fig}
    \vspace{-2ex}
\end{figure}

Figure \ref{fig:grid_search_fig} illustrates the output from the reorientation part of the grid search optimization using motion primitives. As this is a multi-dimensional problem, for visualization purposes  Figure \ref{fig:grid_search_fig} presents only the results when varying the three parameters of the core body lengths. Our prior work in~\cite{olsen2023design} provides more detailed insights with respect to the individual leg optimization for jumping while the link length values are here slightly adjusted as a) we simultaneously account also for attitude maneuvering and b) we consider the complete optimization of leg and torso designs. Figure~\ref{fig:grid_search_fig} demonstrates the performance correlation between body size parameters ($l_{body}$, $w_{body_f}$, and $w_{body_b}$) and the robot's ability to reorient in roll, pitch, and yaw using predefined motion primitives, while maintaining fixed link lengths. The selected robot parameters resulting from this optimization process are listed in Table \ref{tab:OLYMPUSsimulationparameters} and were used for the construction of Olympus. The final robot can be seen in Figure \ref{fig:front_figure}, the final weight of the robot was $13.8$~\textrm{kg}. For the purposes of the reorientation tests, the max torque was saturated to $8$~\textrm{Nm}, and the springs were not actuated.

\section{RL FOR ATTIUDE CONTROL}\label{sec:RL_attitude}
This section presents the reinforcement learning problem formulation, the simulation setup, the reward function shaping, and steps taken to reduce the sim2real gap. 

\subsection{Problem Formulation}

Let $\mathcal{I}$ represent an inertial fixed-frame, and $\mathcal{B}$ be the robot's body fixed-frame. For any orientation quaternion $\mathbf{q}_\mathcal{B}^{\mathcal{I}}$ of the robot's body expressed in $\mathcal{I}$ and any reference quaternion $\mathbf{q}_\mathcal{R}^{\mathcal{I}}$ we seek to develop a control policy $\boldsymbol{\pi}_C$ that manipulates the vector of motor commands $\theta_r \in \mathbb{R}^{12}$ such that $\mathbf{q}_\mathcal{B}^{\mathcal{I}} \rightarrow \mathbf{q}_\mathcal{R}^{\mathcal{I}}$ (i.e., the current orientation converges to the desired values). As observation space, we consider the quaternion error $\mathbf{q}_\mathcal{B}^{\mathcal{R}} = (\mathbf{q}_\mathcal{R}^\mathcal{I})^{*}\otimes \mathbf{q}_{\mathcal{B}}^\mathcal{I}$, its angular velocity $\boldsymbol{\omega}_\mathcal{B}^\mathcal{R}$, as well as the current motor angles and their angular velocities $\boldsymbol{\theta}_m, \dot{\boldsymbol{\theta}}_m$ respectively. Accordingly, the observation vector takes the form
\vspace{-1ex}
\begin{equation}
    \mathbf{o} = [\mathbf{q}_\mathcal{B}^{\mathcal{R}}~\boldsymbol{\omega}_\mathcal{B}^\mathcal{R}~\boldsymbol{\theta}_m~\dot{\boldsymbol{\theta}}_m]. 
\end{equation}

We consider a low-level motor controller for each of the actuated joints which tracks the commanded joint position. Thus, the action $\mathbf{a} \in \mathbb{R}^{12}$ produced by the policy $\boldsymbol{\pi}_C$ 
that after applying clipping to $[-1,1]$ and linear mapping to the motor limits $[ \boldsymbol{\theta}_r^{\min}, \boldsymbol{\theta}_r^{\max}]$ allows to derive the motor commands
\vspace{-1ex}
\begin{equation}
    \boldsymbol{\theta}_r = g(\mathbf{a}),~g:\textrm{clip}\circ\textrm{linear mapping}, 
\end{equation}
where $\circ$ operator denotes the function composition and $\boldsymbol{\theta}_r \in [ \boldsymbol{\theta}_r^{\min}, \boldsymbol{\theta}_r^{\max}]$ is fed to the respective motor controllers, considering the physical limits of the joint.

\subsection{Policy Learning and Reward Function}

In this work, the control policy $\boldsymbol{\pi}_C$ is identified through reinforcement learning and specifically by applying the \ac{ppo} algorithm. We design the reward function in order to control the torso orientation as follows:
\vspace{-1ex}
\begin{equation}
     r = r_{q} + r_{\omega} + p,
\end{equation}
where the reward terms $r_q,~r_\omega$ are the rewards for the re-orientation and body angular velocity respectively. We reset the episode if physical contact between two links occurs. Then, at the end of the episode, the agent receives the penalty $p = -15.0$. Otherwise, the episode length is defined as $6$~sec to provide enough time for re-orientation and stabilize in the desired orientation. Subsequently, the reward terms read:

\small
\vspace{-1ex}
\begin{align}
    r_{q} &= \nu_q \exp{(-\mu_q(\theta_\mathcal{B}^{\mathcal{R}})^2)} \\
    r_{\omega} &= \left\{ 
    \begin{array}{ c l }
    -\nu_\omega \exp{(-\mu_\omega||\boldsymbol{\omega}_\mathcal{B}^\mathcal{R}||^2)} + 1   & \quad \textrm{if } |\theta_\mathcal{B}^{\mathcal{R}}| > 0.1  \\
    10 & \quad \textrm{otherwise,} 
    \end{array} \right.
\vspace{-2ex}
\end{align}
\normalsize
where the parameters $\nu_*,\mu_*>0$ represent tuning parameters and the angle error reads: $\theta_\mathcal{B}^{\mathcal{R}} = 2\arccos{(\textrm{Re}(\mathbf{q}_\mathcal{B}^{\mathcal{R}})})$ with real part of the quaternion defined as $\textrm{Re}$.
The $r_{q}$ reward allows to achieving orientation tracking performance, while the $r_{\omega}$ encourages initial exploration in the joints' space.

\subsection{Neural Architecture}

\begin{figure}
    \centering
    \includegraphics[width=0.99\linewidth]{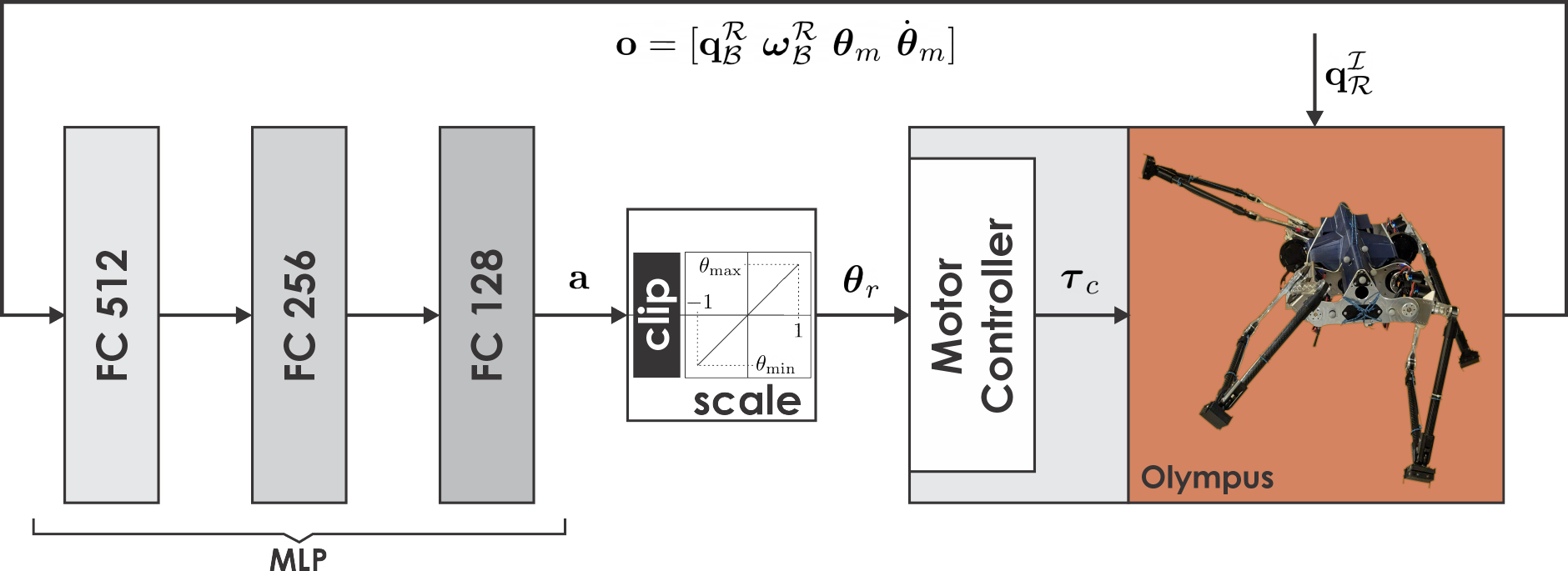}
    \vspace{-4ex}
    \caption{Reinforcement learning controller architecture employed for the problem of in-flight attitude control of the Olympus quadruped.}
    \label{fig:olympus_network}
    \vspace{-4ex}
\end{figure}

The proposed reinforcement learning network consists of three fully-connected layers forming Multilayer perceptron (MLP) with layer widths equal to $[512,256,128]$ and Exponential Linear Unit (ELU) activation, as shown in Figure \ref{fig:olympus_network}.
At the network's output we acquire the estimate of the value function (forming the critic), as well as the mean and standard deviation $\boldsymbol{\mu},\boldsymbol{\sigma}$ of the actor's Gaussian distribution. The mean value of the actor is then clipped to normalized values $[-1,1]$ and re-mapped to the joint angle limits $[ \boldsymbol{\theta}_r^{\min}, \boldsymbol{\theta}_r^{\max}]$ of each of Olympus' motors. For Olympus, the joint angle limits were set to $[-30,125]~\textrm{deg}$ for all motors. 

\begin{figure*}
    \centering
    \subfloat[][Mounted on the rigid rod test platform.]{\includegraphics[width=0.32\linewidth]{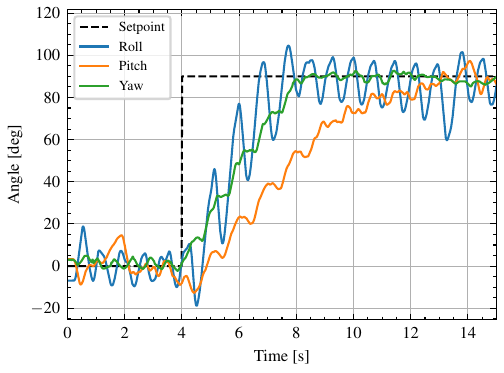}\label{fig:experiments:rollPitchYaw_rod_step}}\hfill
    \subfloat[][Mounted on the rigid rod test platform with varying paw weights with changing setpoint.]{\includegraphics[width=0.32\linewidth]{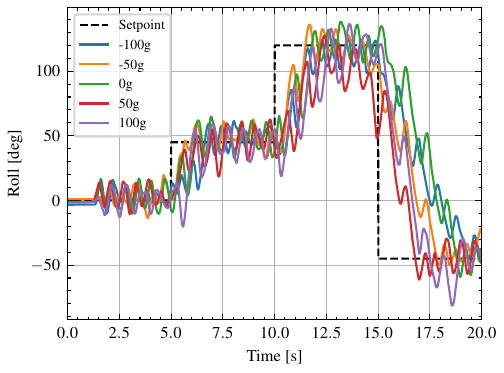}\label{fig:experiments:roll_rod_staircase_all}}\hfill
    \subfloat[][Suspended in the air by rope.]{\includegraphics[width=0.32\linewidth]{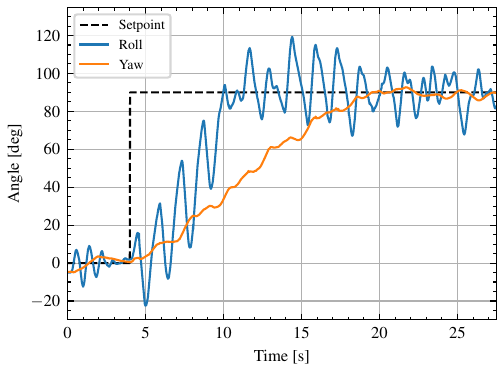}\label{fig:experiments:rollYaw_rope_step}}
    \caption{Closed-loop performance of the RL policy to step inputs in either roll, pitch, or yaw for different setup configurations.}
    \vspace{-4ex}
\end{figure*}
\subsection{Implementation and Crossing the sim2real Gap}

Training of the policy takes place in NVIDIA Isaac Sim~\cite{nvidia_isaac_sim,omniisaacgymenvs2023}, which provides massive parallelization capabilities. 
We run the training of Olympus across $4096$ environments simultaneously on an NVIDIA RTX A6000. The RL Games library~\cite{rl-games2021} is employed and provides the core implementation for \ac{ppo} algorithm.

Transfer of the learned policy to the real robot relies on successfully closing the sim2real gap. To that end, two paths are employed namely a) bringing simulation closer to the true properties of the dynamical properties of the considered quadruped, and b) considering noise and parameter uncertainty at training time. For the first aspect, it is noted that Isaac Sim simulates through a reduced-coordinate representation involving multiple rigid bodies as articulations. The coordinate representation is determined by a root body and accordingly joint angles to all bodies. In this architecture, closed kinematic chains are not supported which implied that the $5$-bar leg design of Olympus had to be kinematically closed through a constraint attached to the ankle joint. This in turn gives rise to stiff differential equations which further implies the need for short simulation step times. Fine-tuning of these parameters relevant to the closed kinematic chain of the real system is the first key step towards successfully closing the sim2real gap. In practice, this implied that the simulator has to run at least at $400~\textrm{Hz}$. Next, to obtain smooth actions and best performance with the motor controllers (Section~\ref{sec:electricalsys}), as well as ensure short training time (around $2\textrm{h}$), we perform inference at $50~\textrm{Hz}$.

A further key step is to identify a faithful model of the motor dynamics. Given Isaac Sim's method to model the open-loop motor dynamics as mere gains, the approach employed involved identifying the values of a PD controller ($K_P^M,K_D^M$) such that the closed-loop angle tracking response in simulation matches that in reality. To that end, MATLAB's system identification tools are employed while explicitly taking into consideration the actuator constraints of the utilized motors. Following on the work in~\cite{elflight}, the resulting simplified motor model is then given by
\vspace{-1ex}
\begin{align}
    \tau = \textrm{sat}(K_P^M(\theta_r-\theta)+K_D^M(0-\dot\theta),\tau_{\max}),
    \vspace{-1ex}
\end{align}
where $\tau_{\max}$ is the motor torque limits, and $\theta$ is a motor's angle. As Olympus employs two distinct types of motors, two such models were identified. Additionally, it is noted that accurate CAD modeling allowed precise prediction of the robot's inertia.

Beyond increasing the modeling fidelity as a means to close the sim2real gap, domain randomization is further employed. To that end, during training the agent is made to experience Gaussian noise on the base orientation ($\mathcal{N}(0,3)~\textrm{deg}$), base angular rate ($\mathcal{N}(0,10)~\textrm{deg/sec}$), the joint positions ($\mathcal{N}(0,3)~\textrm{deg}$), and the joint velocities ($\mathcal{N}(0,20)~\textrm{deg/sec}$). Likewise the control gains for the motor PD loop are also randomized as compared to the values obtained from system identification by as much as $\pm 50\%$. Finally, as a further means to ensure robust transfer to reality, the robot is re-spawned in virtually all possible orientations of the torso and all possible leg joint positions, with zero torso's angular and joint velocities, to allow the agent to be trained for the complete envelope of the system.

\section{EVALUATION STUDIES}\label{sec:_Eval_Intro}

\subsection{Simulation Studies}

\begin{figure}[t]
    \centering
    \includegraphics[width=0.95\linewidth]{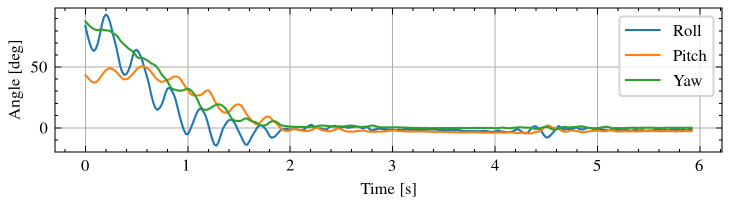}
    \vspace{-2ex}
    \caption{Attitude control performance of the RL policy in simulation performing a $3\textrm{D}$ maneuver driving initial attitude conditions to zero in all roll, pitch, and yaw angles.}
    \label{fig:simulation:control}
    \vspace{-5ex}
\end{figure}

The performance of the attitude re-orientation was first evaluated in simulation. An example of such an evaluation can be seen in Figure~\ref{fig:simulation:control}, where the platform starts with large errors in attitude, and then promptly acts to converge back to the target setpoint of zero in roll, pitch, and yaw.

\subsection{Experiments}\label{sec:experiments}

To assess the platform's reorientation capabilities, two experimental setups were realized: 1. A rotating rigid rod, and 2. Fixed-rope suspension. Roll, pitch, and yaw reorientation tests were conducted on the rotating rod, and roll and yaw were performed on the rope, enabling a comprehensive evaluation of the platform's attitude control performance under varied kinematic constraints, of these tests roll rod, pitch rod, and yaw rope are illustrated in Figure~\ref{fig:front_figure}. In addition, the robot was tested with unmodeled extra masses on the paws. For the control, we utilize the proposed RL framework from Section~\ref{sec:RL_attitude}, which runs on the NVIDIA Jetson Orix NX without any optimizations for faster inference, at an inference rate of $50$~\textrm{Hz}.

\subsubsection{Rotating Rigid Rod}
The platform is rigidly mounted to a testing platform allowing it to spin freely about a single degree of freedom. This setup allows for safe testing of the reorientation capability of the platform about the body-frame $x$ and $y$ and $z$ axes, separately.

First, step inputs are given to the closed-loop system while mounted in configurations isolating roll, pitch, and yaw separately. The results are shown in Figure~\ref{fig:experiments:rollPitchYaw_rod_step}. For the rod test, the robot's workspace intersected partially with the mounting rod, limiting maneuverability. During the pitch test, this more significantly restricted the motors' allowable motion space. This necessitated action clipping to prevent repeated collisions with the mounting rod, consequently limiting pitch reorientation performance by clipping two motors' workspace to only [$-30$,$40$]~\textrm{deg}. While roll and yaw were less affected, the pitch was significantly impacted by gravity, a factor absent in simulations. Hip motors, tasked with maintaining leg positions against gravity, faced scenarios not encountered during policy training, potentially compromising performance.

The design of the paw, and leg as a whole, can have a significant impact on the tracking performance, as the inertia of these elements impacts the effectiveness of the reorientation motions. As such, another experiment is conducted with the same policy, where weights are added and removed from the paws to visualize this effect. This experiment is visualized in Figure~\ref{fig:experiments:roll_rod_staircase_all}, for different weight configurations in the roll setup. From this figure, one can see that by increasing the mass of the paw, thereby increasing the inertia, the platform gains increased attitude maneuvering authority. Therefore, the RL policy becomes faster to converge following the step input. The default paw masses are with a $100~\textrm{g}$ extra mass, representing the mass of a larger paw design more suited for planetary exploration.

\subsubsection{Fixed rope suspension}
In addition to the rigid rod experiments, investigations are done on the platform's re-orientation ability while suspended from a rope to a rotating swivel. While the mounting can prioritize rotation about a certain axis, the remaining degrees of freedom are not as fixed as previously with the rod. As a result, the platform has a greater ability to move in this setup. Another benefit of this mounting is there are fewer restrictions in the workspace of the robot, while at the same time adding very little mass and inertia to the system. The results of providing step inputs to roll and yaw can be seen in Figure~\ref{fig:experiments:rollYaw_rope_step}. Similar to the experiment visualized in Figure~\ref{fig:experiments:rollPitchYaw_rod_step}, the platform is able to successfully re-orient although now with greater oscillations in other axes than the axis being evaluated. This increased oscillation seems to be resulting from the increased mobility available to the platform, and the fact that it can now swing in the rope fixture. This is contrary to the rod setup, where the platform's position was fixed at all times and was only able to rotate about a single axis.

\section{CONCLUSIONS AND FUTURE WORK}\label{sec:conclusions}

This paper presented Olympus, a legged system optimized for dynamic locomotion and in-flight reorientation in lower-gravity environments. Tested in rod-mounted and rope-suspended scenarios, Olympus demonstrated effective reorientation capabilities using a simulation-trained policy. The robot achieved 90-degree step inputs across all axes, adapting to unmodeled masses and varying setpoints. Simulation tests further validated its ability to correct combined roll, pitch, and yaw offsets. Olympus's robust design and capabilities make it suitable for exploring challenging terrains on other planetary bodies, such as Martian lava tubes. Future work will focus on reducing the sim-to-real gap, enhancing reorientation speed, and testing the system's jumping capabilities.

\bibliographystyle{IEEEtran}

\bibliography{./ICRA2025.bib}

\end{document}